\pdfoutput=1

\documentclass[11pt,a4paper]{article}
\usepackage[hyperref]{acl2020}
\usepackage{times}
\usepackage{latexsym}

\usepackage{graphicx}
\usepackage{multirow}
\usepackage{multicol}

\usepackage{microtype}

\aclfinalcopy
\def\aclpaperid{1529}

\title{Dynamically Adjusting Transformer Batch Size by\\ Monitoring Gradient Direction Change}

\author{
Hongfei Xu$^{1,2}$\ \ \ \ Josef van Genabith$^{1,2}$\ \ \ \ Deyi Xiong$^{3}$\ \ \ \ Qiuhui Liu$^4$\thanks{\ \ \ \ Corresponding author.}\\
$^1$Saarland University / Saarland, Germany\\
$^2$German Research Center for Artificial Intelligence / Saarland, Germany\\
$^3$Tianjin University / Tianjin, China\\
$^4$China Mobile Online Services / Henan, China\\
hfxunlp@foxmail.com,
Josef.Van\_Genabith@dfki.de,\\
dyxiong@tju.edu.cn,
liuqhano@foxmail.com
}
\date{}

\begin{document}
\maketitle
\begin{abstract}
The choice of hyper-parameters affects the performance of neural models. While much previous research \cite{Sutskever2013on,John2011Adaptive,Kingma15Adam} focuses on accelerating convergence and reducing the effects of the learning rate, comparatively few papers concentrate on the effect of batch size. In this paper, we analyze how increasing batch size affects gradient direction, and propose to evaluate the stability of gradients with their angle change. Based on our observations, the angle change of gradient direction first tends to stabilize (i.e. gradually decrease) while accumulating mini-batches, and then starts to fluctuate. We propose to automatically and dynamically determine batch sizes by accumulating gradients of mini-batches and performing an optimization step at just the time when the direction of gradients starts to fluctuate. To improve the efficiency of our approach for large models, we propose a sampling approach to select gradients of parameters sensitive to the batch size. Our approach dynamically determines proper and efficient batch sizes during training. In our experiments on the WMT 14 English to German and English to French tasks, our approach improves the Transformer with a fixed $25k$ batch size by $+0.73$ and $+0.82$ BLEU respectively.
\end{abstract}

\section{Introduction}

The performance of neural models is likely to be affected by the choice of hyper-parameters. While much previous research \cite{Sutskever2013on,John2011Adaptive,Kingma15Adam} focuses on accelerating convergence and reducing the effects of the learning rate, comparatively few papers concentrate on the effect of batch size.

However, batch size is also an important hyper-parameter, and some batch sizes empirically lead to better performance than the others.

Specifically, it has been shown that the performance of the Transformer model \cite{vaswani2017attention} for Neural Machine Translation (NMT) \cite{bahdanau2014neural,gehring2017convolutional,vaswani2017attention} relies heavily on the batch size \cite{Martin2018Training,ott2018scaling,abdou2017variable,zhang2019improving}.

The influence of batch size on performance raises the question, how to dynamically find proper and efficient batch sizes during training? In this paper, we investigate the relationship between the batch size and gradients, and propose a dynamic batch size approach by monitoring gradient direction changes. Our contributions are as follows:

\begin{table*}[t]
	\centering
	\begin{tabular}{|l|r|r|r|r|r|r|r|r|r|r|}
		\hline
		k & 1     & 2     & 3     & 4     & 5     & 6     & 7     & 8     & 9     & 10 \\
		\hline
		Size & 4064 & 8994 & 12768 & 17105 & 21265 & 25571 & 29411 & 33947 & 38429 & 43412 \\
		\hline
		$a(g_0^{k - 1},g_0^k)$ &       & 51.52 & 30.37 & 27.42 & 22.61 & 20.87 & 19.80 & 19.59 & 18.92 & 19.23 \\
		\hline
		$a(g_0^{k - 3},g_0^k)$ &       &       &       & 59.53 & 44.20 & 41.77 & 35.34 & 32.19 & 32.10 & 34.29 \\
		\hline
	\end{tabular}
	\caption{The direction change of gradients while accumulating mini-batches.}
	\label{tab:anglenbsize}
\end{table*}

\begin{itemize}
\item We observe the effects on gradients with increasing batch size, and find that a large batch size stabilizes the direction of gradients;
\item We propose to automatically determine dynamic batch sizes in training by monitoring the gradient direction change while accumulating gradients of small batches;
\item To measure gradient direction change efficiently with large models, we propose an approach to dynamically select those gradients of parameters/layers which are sensitive to the batch size;
\item In machine translation experiments, our approach improves the training efficiency and the performance of the Transformer model.
\end{itemize}

\section{Gradient Direction Change and Automated Batch Size}

Gradients indicate the direction and size of parameter updates to minimize the loss function in training. To reveal the effects of the batch size in optimization, we evaluate its influence on the direction change of gradients.

\subsection{Gradient Direction Change with Increasing Batch Size}

To investigate the influence of batch size on gradient direction, we gradually accumulate gradients of small mini-batches as the gradients of a large batch that consists of those mini-batches, and observe how the direction of gradients varies.

Let $d_i^j:(x_i^j, y_i^j)$ stands for the large batch concatenated from the $i$th mini-batch to the $j$th mini-batch, where $x_i^j$ and $y_i^j$ are inputs and targets. Then the gradients $g_i^j$ of model parameters $\theta$ on $d_i^j$ are:

\begin{equation}
	g_i^j = \frac{{\partial L(\theta ,x_i^j,y_i^j)}}{{\partial \theta }}
	\label{eqa:gbij}
\end{equation}

In gradient accumulation, the gradients $g_0^k$ are the sum of $g_0^{k - 1}$ and $g_k^k$:

\begin{equation}
	g_0^k = g_0^{k - 1} + g_k^k
	\label{eqa:acgk}
\end{equation}

To measure the change of gradient direction during accumulation, we regard the two gradients $g_0^{k - 1}$ and $g_0^k$ as $2$ vectors, and compute the angle $a(g_0^{k - 1},g_0^k)$ between them:

\begin{equation}
	a(g_0^{k - 1},g_0^k) = \arccos (\frac{{g_0^{k - 1}  \bullet g_0^k}}{{|g_0^{k - 1}||g_0^k |}})
	\label{eqa:angle}
\end{equation}

\noindent where ``$\bullet$'' indicates inner-product of vectors.

We use the angle of $2$ vectors rather than cosine similarity because:

\begin{itemize}
	\item The angle indicates the change between gradient directions;
	\item When the angle is small, a significant change in the angle only results in a subtle difference in cosine similarity.\footnote{ $cos(5^\circ ) \approx 0.9961$, $cos(10^\circ ) \approx 0.9848$.}
\end{itemize}

We observe the gradient direction varying during accumulating gradients of a Transformer model training on the WMT 14 English-German task following the setting of \newcite{vaswani2017attention} with a batch size of around $50k$ target tokens. To achieve the gradient of the large batch size, we gradually accumulate gradients of mini-batches with around $4k$ target tokens.

Table \ref{tab:anglenbsize} shows a typical example: (i) gradient change is high at the beginning, (ii) gradient change reduces with increasing batch size and (iii) eventually it will start fluctuating (here at k=10).\footnote{By comparing $\sum\limits_{i = 0}^n {a(g_0^{k - i - 1},g_0^{k - i})}$ with $a(g_0^{k - n - 1},g_0^k)$, we can find the direction changes from $g_0^{k - i - 1}$ to $g_0^k$ are inconsistent. Otherwise, $\sum\limits_{i = 0}^n {a(g_0^{k - i - 1},g_0^{k - i})}  \approx a(g_0^{k - n - 1},g_0^k)$.}

Intuitively, the less the direction of accumulated gradients is moved by the gradients of a new mini-batch, the more certainty there is about the gradient direction. Thus we propose that  the magnitude of the angle fluctuation relates to the certainty of the model parameter optimization direction, and may therefore serve as a measure of optimization difficulty.

\subsection{Automated Batch Size with Gradient Direction Change}

Table \ref{tab:anglenbsize} shows that the optimization direction is less stable with a small batch than with a large batch. But after the direction of gradients has stabilized, accumulating more mini-batches seems useless as the gradient direction starts to fluctuate.

Thus, we suggest to compute dynamic and efficient batch sizes by accumulating gradients of mini-batches, while evaluating the gradient direction change with each new mini-batch, and stop accumulating more mini-batches and perform an optimization step when the gradient direction fluctuates.

In practice, we only monitor $a(g_0^{k - 1},g_0^k)$ for efficiency. We record the minimum angle change $a_{min}$ while accumulating gradients, and suppose the gradient direction starts to fluctuate, stop accumulating more mini-batches when $a(g_0^{k - 1},g_0^k) > a_{min} * \alpha$. In this way we can achieve a dynamic batch size (the size of $d_0^k$), where $\alpha$ is a pre-specified hyper-parameter.

\subsection{Efficiently Monitoring Gradient Direction Change}

In practice, a model may have a large amount of parameters, and the cost of computing the cosine similarity between two corresponding gradient vectors are relatively high. To tackle this issue, we propose to divide model parameters into groups, and monitor gradient direction change only on a selected group in each optimization step. For a multi-layer model, i.e. the Transformer, a group may consist of parameters of $1$ layer or several layers.

To select the parameter group which is sensitive to the batch size, we record the angles of gradient direction change $a(g_0^0,g_0^1), ..., a(g_0^{k - 1},g_0^k)$ in the gradient accumulation, and define $a_{max}$ and $a_{min}$ as the maximum and minimum direction change:

\begin{equation}
	a_{max} = max(a(g_0^0,g_0^1), ..., a(g_0^{k - 1},g_0^k))
	\label{eqa:amax}
	\end{equation}
	\begin{equation}
	a_{min} = min(a(g_0^0,g_0^1), ..., a(g_0^{k - 1},g_0^k))
	\label{eqa:amin}
\end{equation}

We then use $\Delta a$ to measure the uncertainty reduction in the optimization direction:

\begin{equation}
	\Delta a = a_{max} - a_{min}
	\label{eqa:achieve}
\end{equation}

Intuitively, the optimization direction of the parameter group which results in a larger $\Delta a$ profits more from the batch size, and the group with a larger $\Delta a$ should be more frequently sampled.

We average the recent history of $\Delta a_k$ of the $k$th parameter group into $\overline {\Delta {a_k}}$. Inspired by \newcite{gumbel1954statistical,Maddison2014A,zhang2019bridging}, we first add Gumble noise to each $\overline {\Delta {a_k}}$ to prevent the selection falling into a fixed group:

\begin{equation}
	\Delta a_k^* = \overline {\Delta {a_k}} - \log ( - \log u)
	\label{eqa:gumble}
\end{equation}

\noindent where $u \in (0,1)$ is a uniform distribution.

Then we zero negative values\footnote{$\Delta a_k$ is positive, but after adding Gumble noise, there is a small possibility that it turns negative. In our case, negative values only occur very few times.} in $\Delta a_1^*$, ..., $\Delta a_n^*$ and normalize them into a probability distribution:

\begin{equation}
	{p_k} = \frac{{\Delta a{{_k^*}^\beta }}}{{\sum\limits_{i = 1}^n {\Delta a{{_i^*}^\beta }} }}
	\label{eqa:psample}
\end{equation}

We use $p_k$ as the probability to sample the $k$th group, and $\beta$ is a hyper-parameter to sharpen the probability distribution. We do not use softmax because it would heavily sharpen the distribution when the gap between values is large, and makes it almost impossible to select and evaluate the other groups in addition to the one with highest $\Delta a_k^*$.\footnote{For example, the result of softmax over [22, 31, 60] is [3.13e-17, 2.54e-13, 1.00], the last element takes almost all possibility mass. But we later find that if $\Delta a$ is normalized ($\Delta a = (a_{max} - a_{min}) / a_{max}$) in Equation \ref{eqa:achieve}, the softmax works comparably well, which avoids using the hyper parameter $\beta$ in Equation \ref{eqa:psample}.}

\section{Experiments}

\begin{table}[t]
	\centering
	\begin{tabular}{|l|r|r|r|}
		\hline
		Batch Size & \multicolumn{1}{l|}{En-De} & \multicolumn{1}{l|}{En-Fr} & \multicolumn{1}{l|}{Time} \\
		\hline
		25k   & 27.38 & 39.34 & 35h21m \\
		\hline
		50k   & 27.93 & 39.97 & 60h38m \\
		\hline
		dyn  & \textbf{28.11}$^\dag$ & \textbf{40.16}$^\dag$ & \textbf{33h37m} \\
		\hline
	\end{tabular}
	\caption{Performance. Time is the training time on the WMT 14 En-De task for $100k$ training steps. $\dag$ indicates $p<0.01$ in the significance test.}
	\label{tab:bleu}
\end{table}

\begin{table}[t]
	\centering
	\begin{tabular}{|l|r|r|}
		\hline
		& \multicolumn{1}{l|}{En-De} & \multicolumn{1}{l|}{En-Fr} \\
		\hline
		min   & 7069 & 8025 \\
		\hline
		avg   & 26264.19 & 30248.90 \\
		\hline
		max   & 102165 & 103352 \\
		\hline
	\end{tabular}
	\caption{Statistics of Batch Size.}
	\label{tab:bsize}
\end{table}

\begin{figure}[t]
	\centering
	\includegraphics[width=1.0\columnwidth]{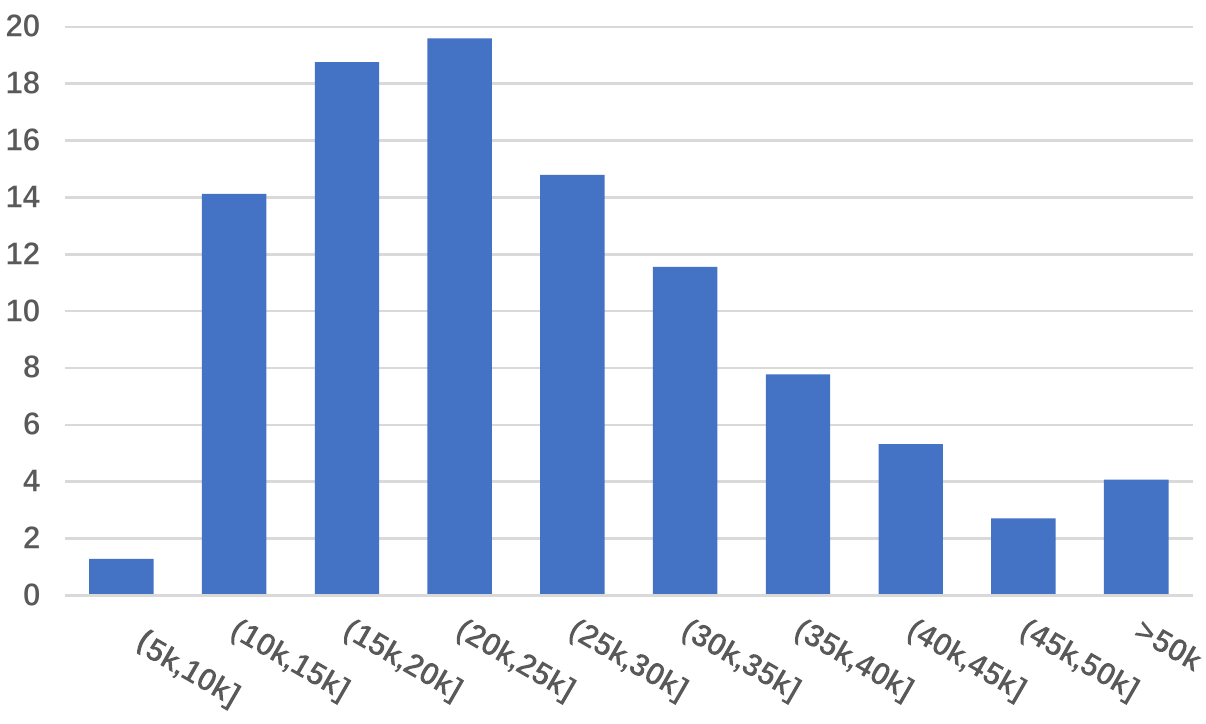}\\
	\caption{Distribution of Dynamic Batch Sizes. Values on y-axis are percentages.}\label{fig:fbsize}
\end{figure}

\begin{figure}[t]
	\centering
	\includegraphics[width=1.0\columnwidth]{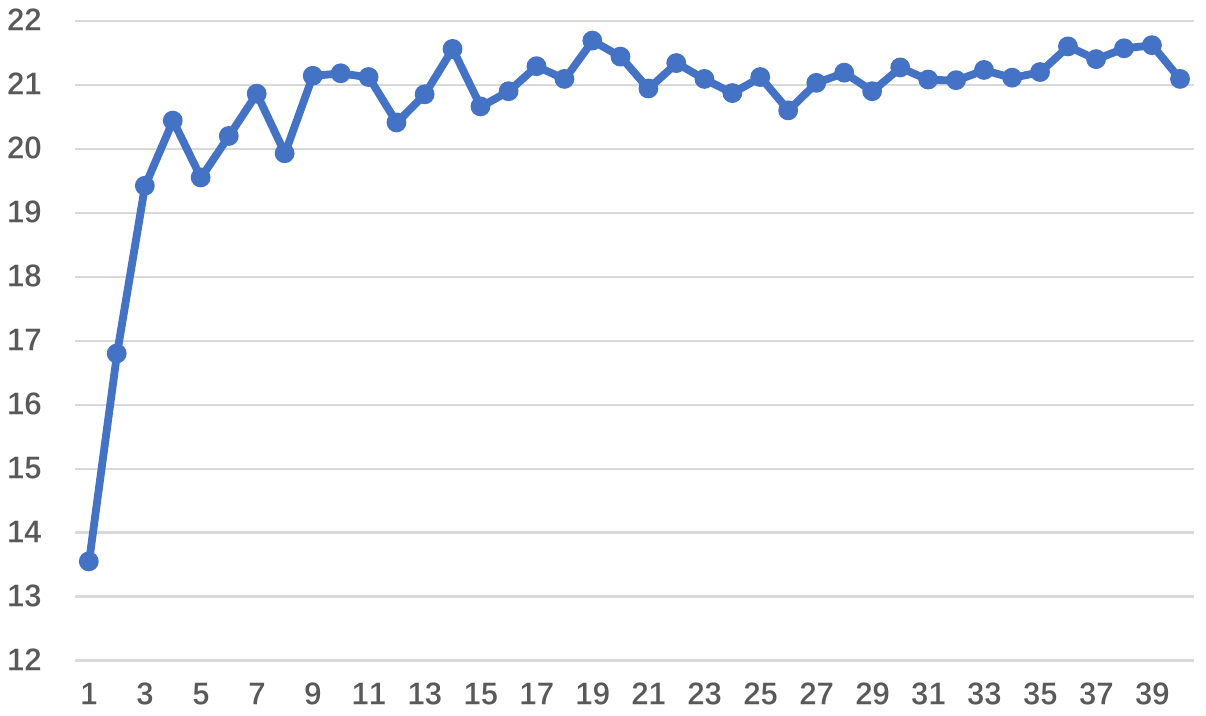}\\
	\caption{Minimum Gradient Direction Change during Training. X-axis 2.5k training steps, y averaged $a_{min}$ (Equation \ref{eqa:amin}).}\label{fig:consis}
\end{figure}

\begin{table}[t]
	\centering
	\begin{tabular}{|r|r|r|r|r|}
		\hline
		\multirow{2}[4]{*}{$\alpha$} & \multicolumn{2}{c|}{Batch Size} & \multicolumn{1}{c|}{\multirow{2}[4]{*}{BLEU}} & \multicolumn{1}{c|}{\multirow{2}[4]{*}{Time}} \\
		\cline{1-3}   & \multicolumn{1}{l|}{avg} & \multicolumn{1}{l|}{max} &       &  \\
		\hline
		1.0   & 19367.76 & 60945 & 27.90 & \textbf{24h50m} \\
		\hline
		1.1   & 26264.19 & 102165 & 28.11 & 33h37m \\
		\hline
		1.2   & 36208.47 & 164908 & \textbf{28.39} & 46h04m \\
		\hline
		1.3   & 51470.34 & 205210 & 28.37 & 63h56m \\
		\hline
	\end{tabular}
	\caption{Effects of Different $\alpha$.}
	\label{tab:alpha}
\end{table}

We implemented our approaches based on the Neutron implementation \citep{xu2019neutron} of the Transformer translation model. We applied our approach to the training of the Transformer, and to compare with \newcite{vaswani2017attention}, we conducted our experiments on the WMT 14 English to German and English to French news translation tasks on $2$ GTX 1080Ti GPUs. Hyper parameters were tuned on the development set (newstest 2012 and 2013). We followed all settings of \newcite{vaswani2017attention} except for the batch size. We used a beam size of $4$ for decoding, and evaluated case-sensitive tokenized BLEU\footnote{\url{https://github.com/moses-smt/mosesdecoder/blob/master/scripts/generic/multi-bleu.perl}} with significance test \cite{koehn2004statistical}.

We used an $\alpha$ of $1.1$ to determine the fluctuation of gradient direction by default. We regarded each encoder/decoder layer as a parameter group, and used a $\beta$ of $3$ for the parameter group selection.

\subsection{Performance}

We compared the results of our dynamic batch size approach to two fixed batch size baselines, the $25k$ batch size is the empirical value of \newcite{vaswani2017attention}, while \newcite{zhang2019improving} investigate $50k$ batch size. Results are shown in Table \ref{tab:bleu} with the statistics of batch sizes of our approach shown in Table \ref{tab:bsize} and the detailed distribution of batch sizes for the En-De task shown in Figure \ref{fig:fbsize}.

Table \ref{tab:bleu} and \ref{tab:bsize} show that our approach outperforms both the fixed $25k$ and $50k$ batch size settings with an average batch size around $26k$, and our approach is slightly faster than the $25k$ setting despite of the additional cost for monitoring gradient direction change.\footnote{It is hard to accumulate an accurate $25k$ target tokens in a batch, and in fact, the fixed $25k$ setting results in an average batch size of $26729.79$.}

Figure \ref{fig:fbsize} shows an interesting fact that the most frequently used automated batch sizes were close to the fixed value ($25k$) of \newcite{vaswani2017attention}.

\subsection{Analysis of Minimum Gradient Direction Change}

In order to observe the varying of minimum gradient direction change during training, we averaged the minimum angle for every $2.5k$ training steps. Results are shown in Figure \ref{fig:consis}.

Figure \ref{fig:consis} shows that the minimum direction change of gradients was small at the beginning, and gradually increased with training. Given that a small angle change indicates that there is more certainty in the gradient direction, this observation is consistent with the fact that finding the optimization direction is harder and harder with training.

\subsection{Effects of $\alpha$}

We studied the effects of different $\alpha$ values on the En-De task, and results are shown in Table \ref{tab:alpha}.\footnote{We observed that the minimum batch size does not change significantly with increasing $\alpha$, so we omit it for space.}

Table \ref{tab:alpha} shows that with increasing $\alpha$, the average batch size and the time cost increases along with the performance. A wide range of values works relatively well indicating that its selection is robust, and $1.1$ seems to be a good trade off between the cost and the performance in our experiments.\footnote{For $\alpha=1.2$ on the En-Fr task, the corresponding values are: 44294.16, 185972, \textbf{40.35} and 54h12m.} It is also worth noting that $\alpha=1$ outperforms the $25k$ baseline while being $1.42$ times faster (Table \ref{tab:bleu}).

\section{Related Work}

\newcite{Martin2018Training} demonstrate that the batch size affects the performance of the Transformer, and a large batch size tends to benefit performance, but they use fixed batch sizes during training. \newcite{abdou2017variable} propose to use a linearly increasing batch size from 65 to 100 which slightly outperforms their baseline. \newcite{smith2018dont} show that the same learning curve on both training and test sets can be obtained by increasing the batch size during training instead of decaying the learning rate.

For fast convergence, \newcite{Balles2017Coupling} propose to approximately estimate the mean value of the batch size for the next batch by maximizing the expected gain with a sample gradient variance ($||g|{|^2}$) computed on the current batch, while our approach compares the gradient direction of change ($a(g_0^{k - 1},g_0^k)$) during accumulation of mini-batches in the assembling of a large batch.

We suggest our approach is complementary to \newcite{Sutskever2013on,John2011Adaptive,Kingma15Adam}, as their approaches decide the magnitude of the move in the optimization direction, while our approach provides reliable gradient direction.

\section{Conclusion}

In this paper, we analyze the effects of accumulated batches on the gradient direction, and propose to achieve efficient automated batch sizes by monitoring change in gradient accumulation and performing an optimization step when the accumulated gradient direction is almost stable. To improve the efficiency of our approach with large models, we propose a sampling approach to select gradients of parameters sensitive to the batch size.

Our approach improves the Transformer with a fixed $25k$ batch size by $+0.73$ and $+0.82$ BLEU on the WMT 14 English to German and English to French tasks respectively while preserving efficiency.

\section*{Acknowledgments}

We thank anonymous reviewers for their insightful comments. Hongfei Xu acknowledges the support of China Scholarship Council ([2018]3101, 201807040056). Deyi Xiong is supported by the National Natural Science Foundation of China (Grant No. 61861130364), the Natural Science Foundation of Tianjin (Grant No. 19JCZDJC31400) and the Royal Society (London) (NAF$\backslash$R1$\backslash$180122). Hongfei Xu and Josef van Genabith are supported by the German Federal Ministry of Education and Research (BMBF) under the funding code 01IW17001 (Deeplee).

\bibliography{acl2020}
\bibliographystyle{acl_natbib}
\end{document}